\documentclass[letterpaper, 10 pt, conference]{ieeeconf}  

\IEEEoverridecommandlockouts                              

\usepackage{amsmath,amssymb,amsfonts}
\usepackage{balance}
\usepackage{cite}
\usepackage{algorithmic}
\usepackage{graphicx}
\usepackage{textcomp}
\usepackage{xcolor}
\usepackage{csquotes}
\usepackage{siunitx}
\usepackage{subcaption}
\usepackage{gensymb}
\usepackage{url}
\usepackage{multirow}
\usepackage{colortbl}
\usepackage{wasysym}
\usepackage{booktabs}

\makeatletter
\let\NAT@parse\undefined
\makeatother
\usepackage{hyperref}
\hypersetup{
    colorlinks,
    linkcolor=[HTML]{B75659},
    citecolor=[HTML]{B75659},
    urlcolor=[HTML]{6595BF}
}

\title{\LARGE \bf
Self-Mixing Laser Interferometry: In Search of an Ambient Noise-Resilient Alternative to Acoustic Sensing
}

\author{Remko Proesmans$^{1}$, Thomas Lips$^{1}$ and Francis wyffels$^{1}$
\thanks{*This work was supported by Research Foundation Flanders (grant no. 1S15925N and 1S56024N), and by euROBIN (grant no. 101070596). }
\thanks{$^{1}$Remko Proesmans, Thomas Lips and Francis wyffels are with the AI and Robotics Lab (IDLab-AIRO), Ghent University---imec, Ghent, Belgium
        {\tt\small remko.proesmans@ugent.be}}%
}

\def\BibTeX{{\rm B\kern-.05em{\sc i\kern-.025em b}\kern-.08em
    T\kern-.1667em\lower.7ex\hbox{E}\kern-.125emX}}
\begin{document}

\maketitle
\thispagestyle{empty}
\pagestyle{empty}

\begin{abstract}

Self-mixing interferometry (SMI) has been lauded for its sensitivity in detecting microvibrations, while requiring no physical contact with its target.
Microvibrations, i.e., sounds, have recently been used as a salient indicator of extrinsic contact in robotic manipulation.
In previous work, we presented a robotic fingertip using SMI for extrinsic contact sensing as an ambient-noise-resilient alternative to acoustic sensing.
Here, we extend the validation experiments to the frequency domain.
We find that for broadband ambient noise, SMI still outperforms acoustic sensing, but the difference is less pronounced than in time-domain analyses.
For targeted noise disturbances, analogous to multiple robots simultaneously collecting data for the same task, SMI is still the clear winner.
Lastly, we show how motor noise affects SMI sensing more so than acoustic sensing, and that a higher SMI readout frequency is important for future work.
Design and data files are available at \href{https://github.com/RemkoPr/icra2025-SMI-tactile-sensing}{https://github.com/RemkoPr/icra2025-SMI-tactile-sensing}.

\end{abstract}

\section{INTRODUCTION}

\begin{figure}
\centering
\includegraphics[width=0.6\linewidth]{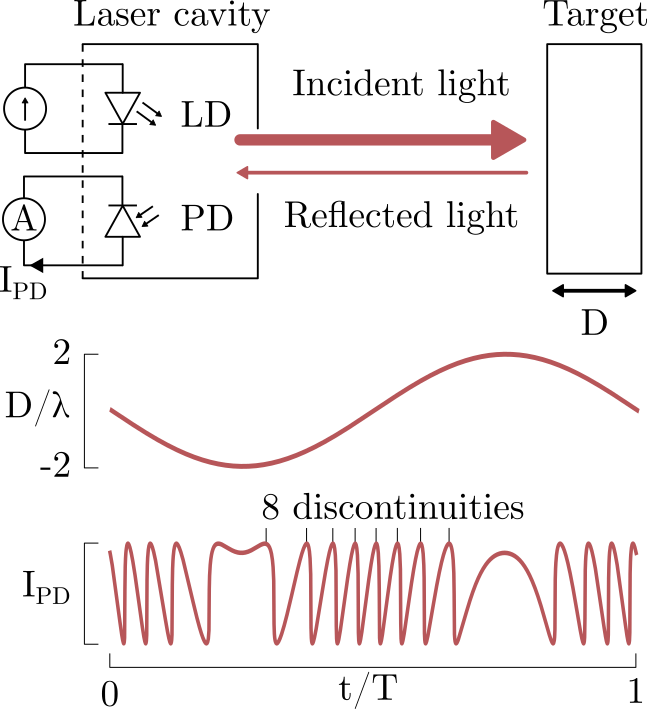}
\caption{Self-mixing interferometry. Every time the target displacement D changes by $\lambda$/2, a discontinuity or fringe appears in the SMI signal. }
\label{fig:smi_principle}
\end{figure}

Self-mixing is a phenomenon where light emitted from a laser diode (LD) is reflected back into the laser cavity and interferes with the resonating light in the cavity~\cite{yu2016}.
A photodiode (PD) in the cavity can measure this interference pattern.
This is called self-mixing interferometry (SMI).
Fig.~\ref{fig:smi_principle} shows an SMI signal simulated based on \cite{plantier2005}.
The characteristic shape of the PD current I$_{\text{PD}}$ is explained as follows: the distance D between the laser and the target determines the phase difference $\phi$ between the resonating light in the cavity and the reflected light, which in turn determines the light power in the cavity. 
When $\Delta$D$=\lambda/2$, the optical path difference for the reflected light is $\lambda$ and $\phi$ is the same.
Hence, a fringe occurs for each $\lambda/2$ that the target travels.

SMI has been widely applied across various fields, including the measurement of mechanical transfer functions of, e.g., car doors and micro-electromechanical systems (MEMS)~\cite{donati2018}, as well as biomedical applications like the non-contact measurement of blood pulsation and respiratory sounds~\cite{donati2014}.
Overall, SMI has been lauded for its highly accurate vibrometric capabilities without the need for mechanical contact with the target~\cite{donati2018, donati2014}.

In robotics, the microvibrations that SMI is so attuned to have been used as markers for slip during robotic manipulation.
Such microvibrations have traditionally been captured using accelerometers, piezoelectric elements and microphones~\cite{kyberd2023, slipreview2020}.
Recently, microphones have also been successfully applied to detect vibrations caused by extrinsic contact during robot manipulation tasks.
This acoustic information was found to ease the learning of
contact-rich robot manipulation skills~\cite{liu2024maniwav}.
Microphones share the characteristic property of SMI systems that no mechanical contact with the measurement target is required.
In the case of tactile sensing, this means that the sensor can be located safe from harm behind the finger surface, without getting in the way of any other potential sensors.
A second analogy is that the microvibrations described in the previous paragraph are nothing but mechanically coupled sound waves.
Hence, microphones and SMI both essentially measure sound.

In previous work~\cite{proesmans2025selfmixinglaserinterferometryrobotic}, we developed and compared two robotic fingertips: one featuring acoustic sensing, the other, SMI sensing.
We found that SMI can detect more subtle events than acoustic sensing and that it offers great resilience against broadband ambient noise.
However, our analysis was limited to the time domain and the robot experiments were mostly static.
In this work, we add onto the time-domain results, but also use spectral data to learn a classification task.
We highlight the potential for SMI sensing as a noise-resilient method to detect extrinsic contact, and indicate remaining design challenges to be solved.


\section{Fingertip Design}

This section recapitulates the electrical and mechanical design of both the SMI and the microphone fingertip, refer to~\cite{proesmans2025selfmixinglaserinterferometryrobotic} for a detailed description. 
The complete design files are available on GitHub.

\subsection{Circuit Design}

\begin{figure}
    \centering
    \vspace{2mm}
    \includegraphics[width=0.9\linewidth]{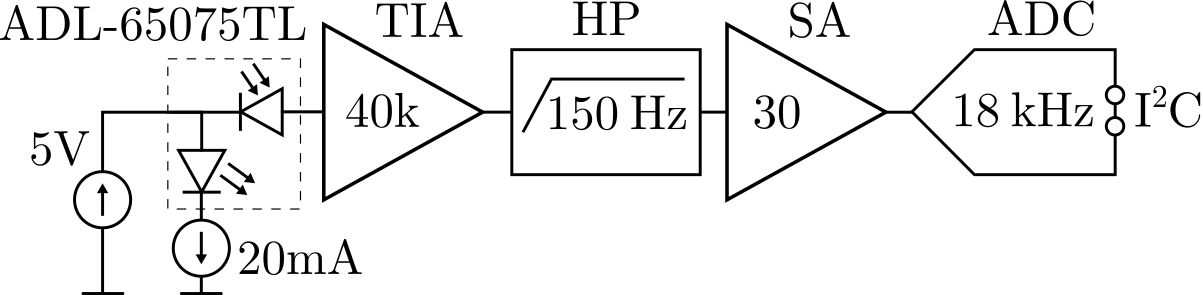} 
    \caption{SMI readout circuit.}
    \label{fig:circuit}
\end{figure}

\subsubsection{SMI sensor} A schematic representation of the SMI circuit is shown in Fig.~\ref{fig:circuit}.
An ADL-65075TL laser ($\lambda$\,=\,650\,nm) with integrated PD is used.
The LD current is fixed at \SI{20}{mA}.
The monitor current of the PD is transduced to a voltage and amplified by a transimpedance amplifier (TIA).
A first-order high-pass (HP) filter rejects the DC component, and the AC component is further amplified by a signal amplifier (SA) with a gain of 30.
Lastly, the signal is sampled by an analog-to-digital converter (ADC) at approximately 18\,kHz.

\subsubsection{Microphone} The microphone is a bottom-port ICS\nobreakdash-43434 with an I2S communication interface on a custom breakout PCB.

\subsection{Mechanical Design} \label{ss:mech}

\begin{figure}[tpb]
\centering
\begin{subfigure}[t]{0.26\textwidth}
    \centering
    \includegraphics[height=3.4cm]{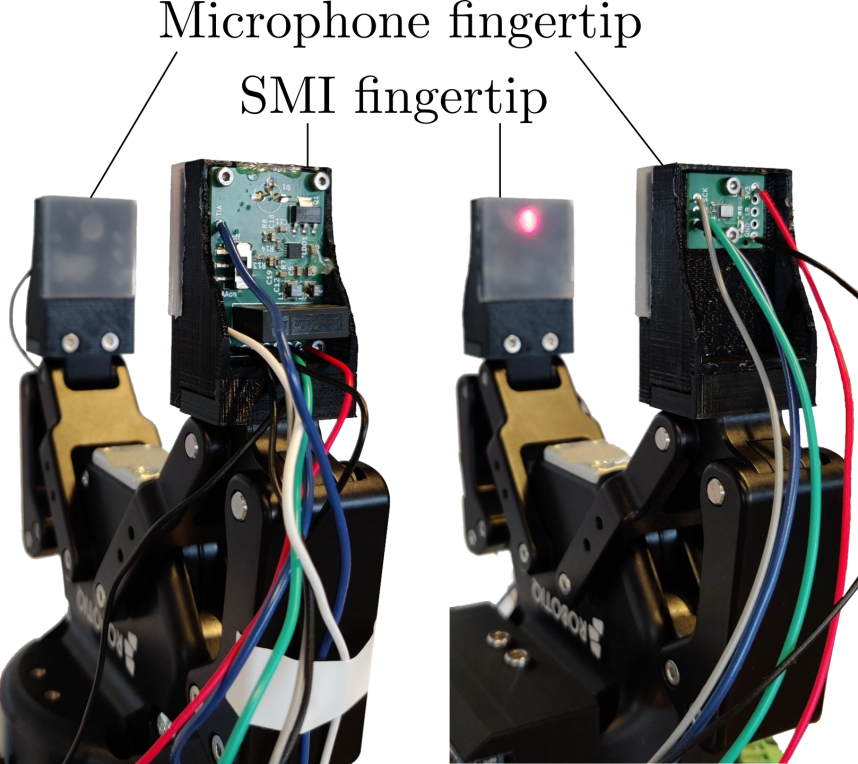}
    \caption{Fingertips mounted on a Robotiq~2F-85 gripper.}
    \label{fig:fingertips}
\end{subfigure}\hfill
\begin{subfigure}[t]{0.2\textwidth}
    \includegraphics[height=3.4cm]{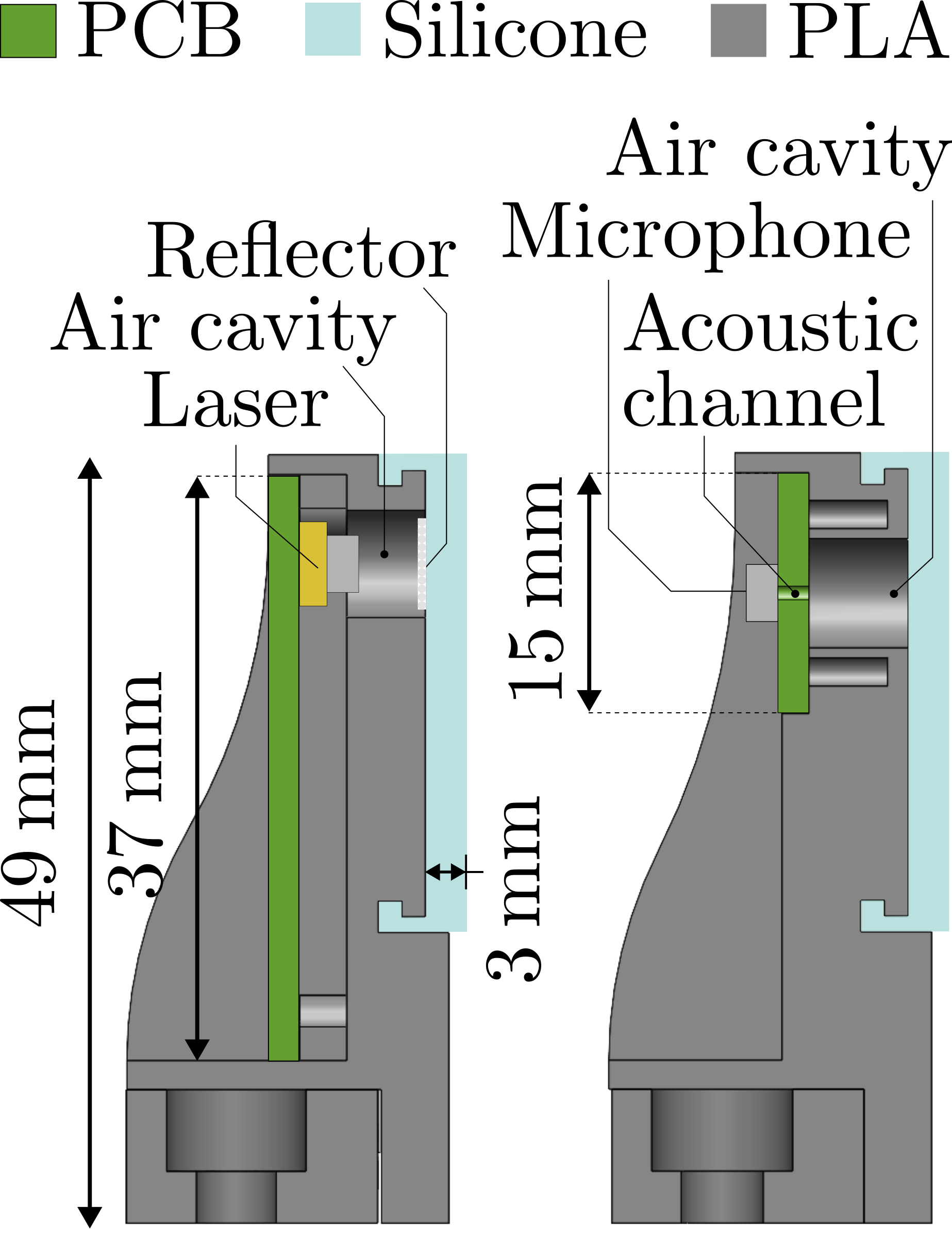} 
    \caption{Section views.}
    \label{fig:section_view}
\end{subfigure}
\caption{Mechanical design of the fingertips.}
\label{fig:mechanical}
\end{figure}

The fingertips consist of 3D printed parts, as well as a 3\,mm thick silicone contact surface, see Fig~\ref{fig:mechanical}.
The laser is facing a piece of retroreflective tape stuck to the back of the silicone contact surface.
The acoustic channel of the bottom-port microphone leads to an enclosed air cavity behind the silicone.

\section{Time-domain Characterisation} \label{s:timedomain}

\begin{figure}[tpb]
\centering
\begin{subfigure}[t]{0.23\textwidth}
    \centering
    \includegraphics[width=0.8\linewidth]{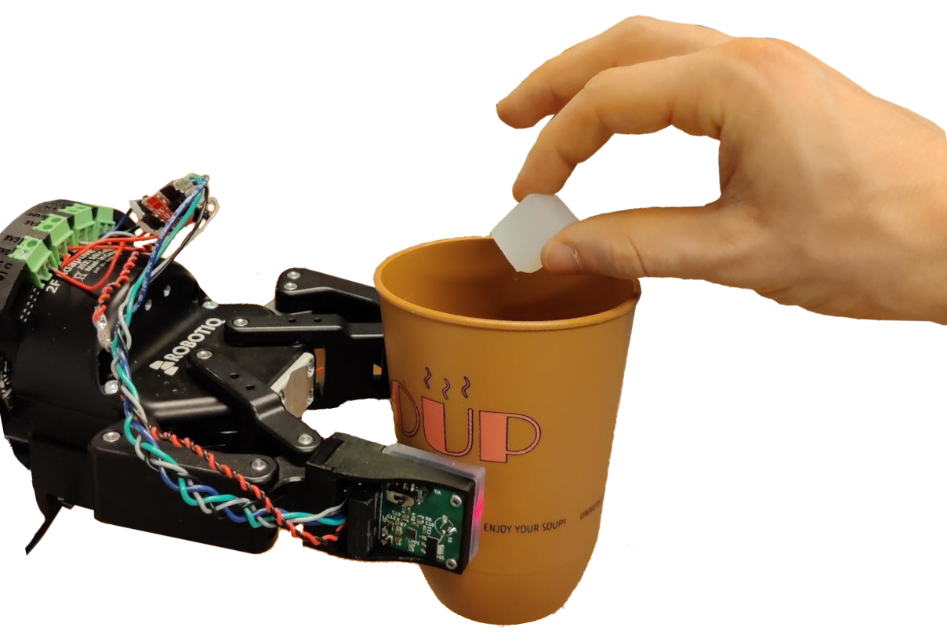} 
    \caption{Pieces of soft silicone are dropped into the cup.}
    \label{fig:cup_silicone_setup}
\end{subfigure}\hfill
\begin{subfigure}[t]{0.23\textwidth}
    \centering
    \includegraphics[width=0.9\linewidth]{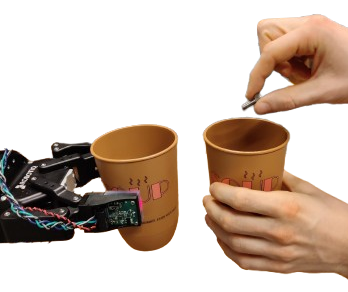} 
    \caption{M6x14 bolts are dropped into the cup held either by the robot or by a person.}
    \label{fig:cup_bolt_setup}
\end{subfigure}

\vspace{2mm}

\begin{subfigure}[t]{0.23\textwidth}
    \centering
    \includegraphics[width=1.05\linewidth]{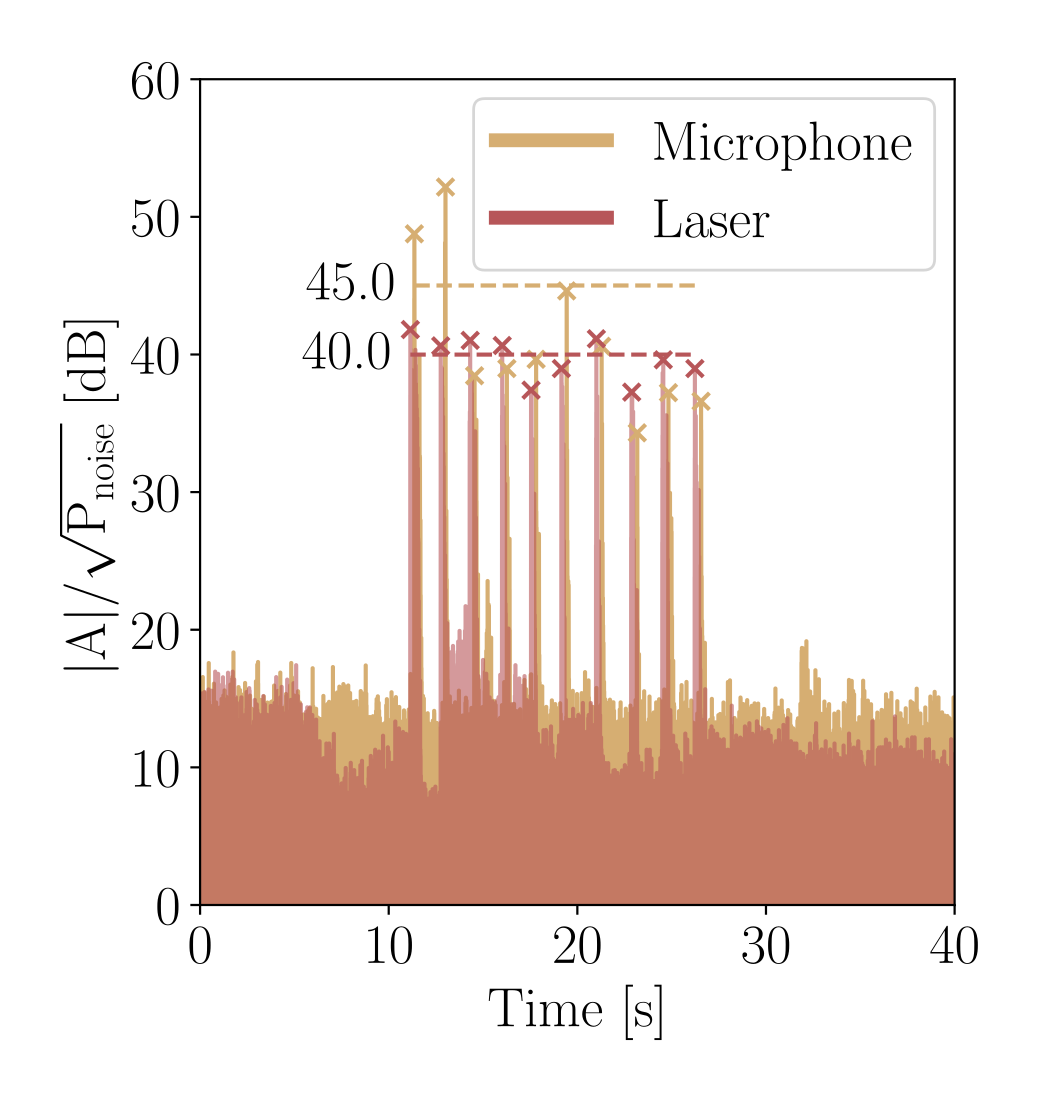} 
    \caption{Silicone: normalised sensor amplitudes, 57\,dB ANL.}
    \label{fig:silicone_plot}
\end{subfigure}\hfill
\begin{subfigure}[t]{0.23\textwidth}
    \centering
    \includegraphics[width=1.05\linewidth]{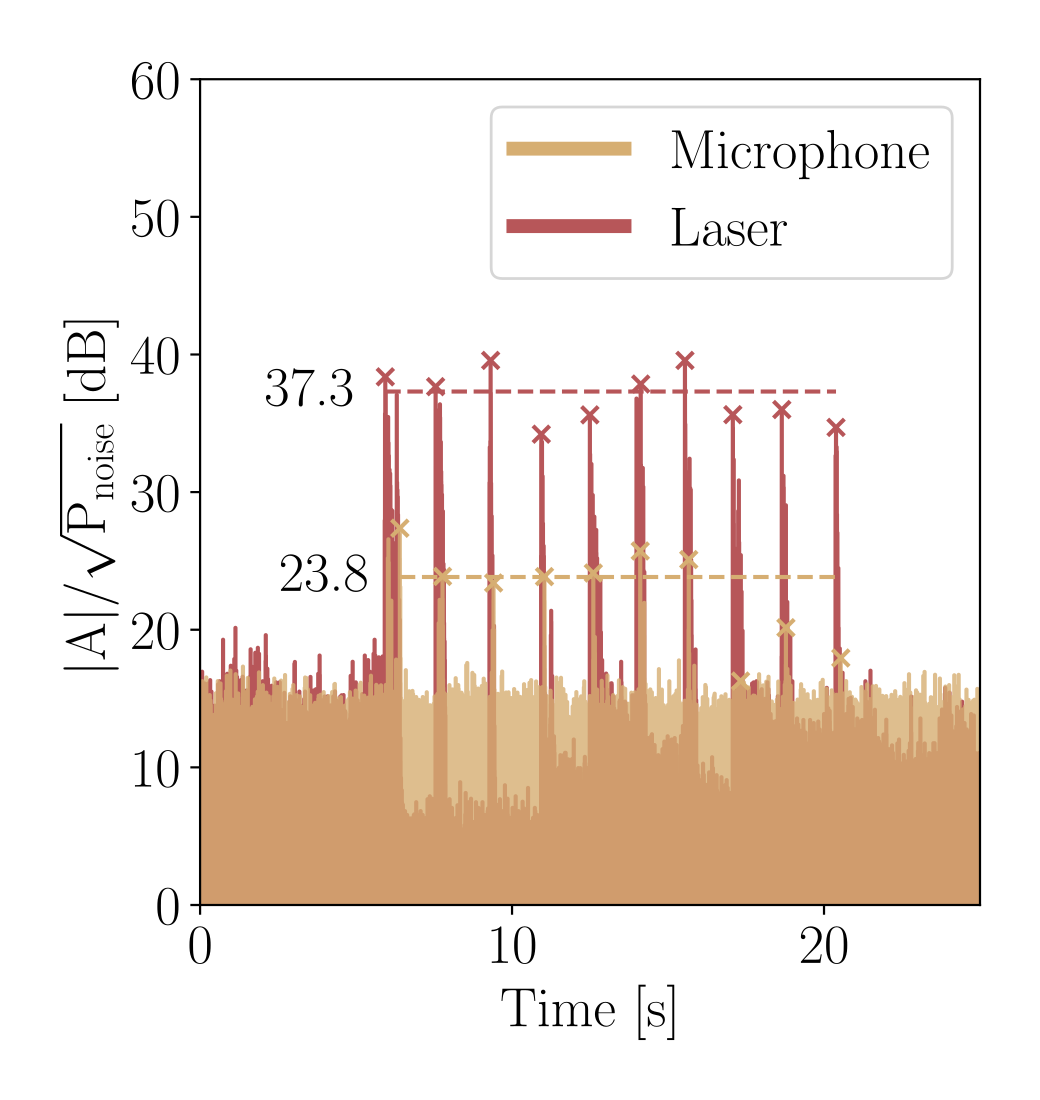} 
    \caption{Silicone: normalised sensor amplitudes, 82\,dB ANL.}
    \label{fig:silicone_noise_plot}
\end{subfigure}

\vspace{2mm}

\begin{subfigure}[t]{0.49\textwidth}
    \centering
    \includegraphics[width=0.8\linewidth]{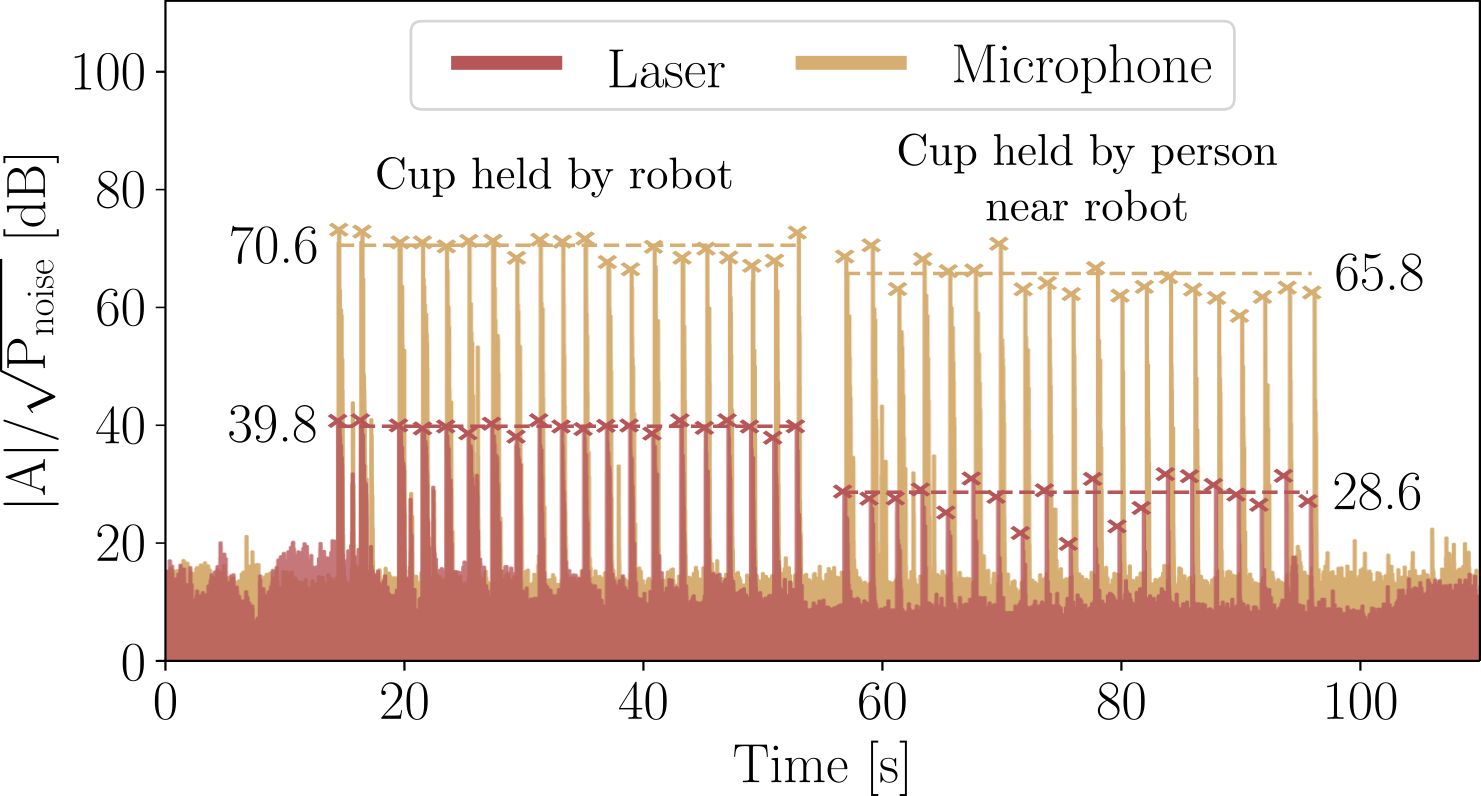} 
    \caption{Bolts: normalised sensor amplitudes, 57\,dB ANL. Left: bolts are dropped into the cup held by the robot. Right: bolts are dropped into a cup held by a person near the robot.}
    \label{fig:bolts_external_cup2}
\end{subfigure}
\caption{Measuring extrinsic contact of objects falling into a hard plastic cup. The laser shows increased isolation from ambient noise.}
\label{fig:cup}
\end{figure}

In this section, we present a time-domain analysis to highlight the potential noise resilience characteristics of SMI and compare to acoustic sensing.

\subsection{Experimental Design}

The robot holds a hard plastic cup, as shown in Fig.~\ref{fig:cup_silicone_setup} and \ref{fig:cup_bolt_setup}. 
First, soft silicone pieces are dropped into the cup at a baseline ambient noise level (ANL) of 57\,dB.
Then, white noise is played from laptop speakers, increasing the ANL to 82\,dB, and the experiment is repeated.
This experiment serves to test sensor resilience to broadband ambient noise.
Second, 14\,mm M6 bolts are dropped into the cup at baseline ANL, after which more bolts are dropped into an identical cup that is held close to the robot.
This way, we test the resilience to \enquote{in-band} ambient noise.

To compare the time-domain sensor output amplitudes $A$, we normalise them with the square root of their noise floor power P$_{\text{noise}}$, i.e. the signal power at rest:
\begin{equation}
    \sqrt{P_{\text{noise}}}=\sqrt{\frac{1}{N}\sum_{i=0}^{N-1} |x_{\text{noise}, i}-\bar{x}_\text{noise}|^2}
\end{equation}
with $N$ the number of measured samples $x_{\text{noise}, i}$, $\bar{x}_\text{noise}$ being the mean of all $x_{\text{noise}, i}$.
The signal-to-noise ratio (SNR) is calculated as:
\begin{equation}
    \text{SNR}=\frac{\sum_{i=1}^{M} |p_i|^2}{P_{\text{noise}}\cdot M}
\end{equation}
with $p_i$ the peak normalised amplitude corresponding to the $i$-th dropped object and $M$ the amount of objects.

\subsection{Results}

When silicone pieces are dropped into the cup, Fig.~\ref{fig:silicone_plot}, the SNR of the microphone is 45\,dB, outperforming the laser by 5\,dB.
With external white noise (Fig.~\ref{fig:silicone_noise_plot}), the laser outperforms the microphone by 13.5\,dB.

In the second experiment, see Fig.~\ref{fig:bolts_external_cup2}, the microphone is much more sensitive than the laser.
However, for the microphone, the difference between bolts dropping into the robot's cup versus the person's cup is 4.8\,dB, compared to 11.2\,dB for the laser: an additional 6.4\,dB \enquote{in-band} noise isolation is gained.
For the microphone, several peaks caused by bolts in the person's cup exceed the lowest peak due to a bolt in the robot's cup: there is no perfect separation between the two cases.
For the laser, there is perfect separation.

\section{Frequency Domain Application} \label{s:experiments}

In this section, we use spectral data to solve a task. 
We are inspired by one of the experiments from ManiWAV\cite{liu2024maniwav} where audio cues are used to guess whether or not a cup is empty while shaking it with a robot arm. 

\subsection{Experimental design}
\begin{figure}
\centering
\includegraphics[width=0.6\linewidth]{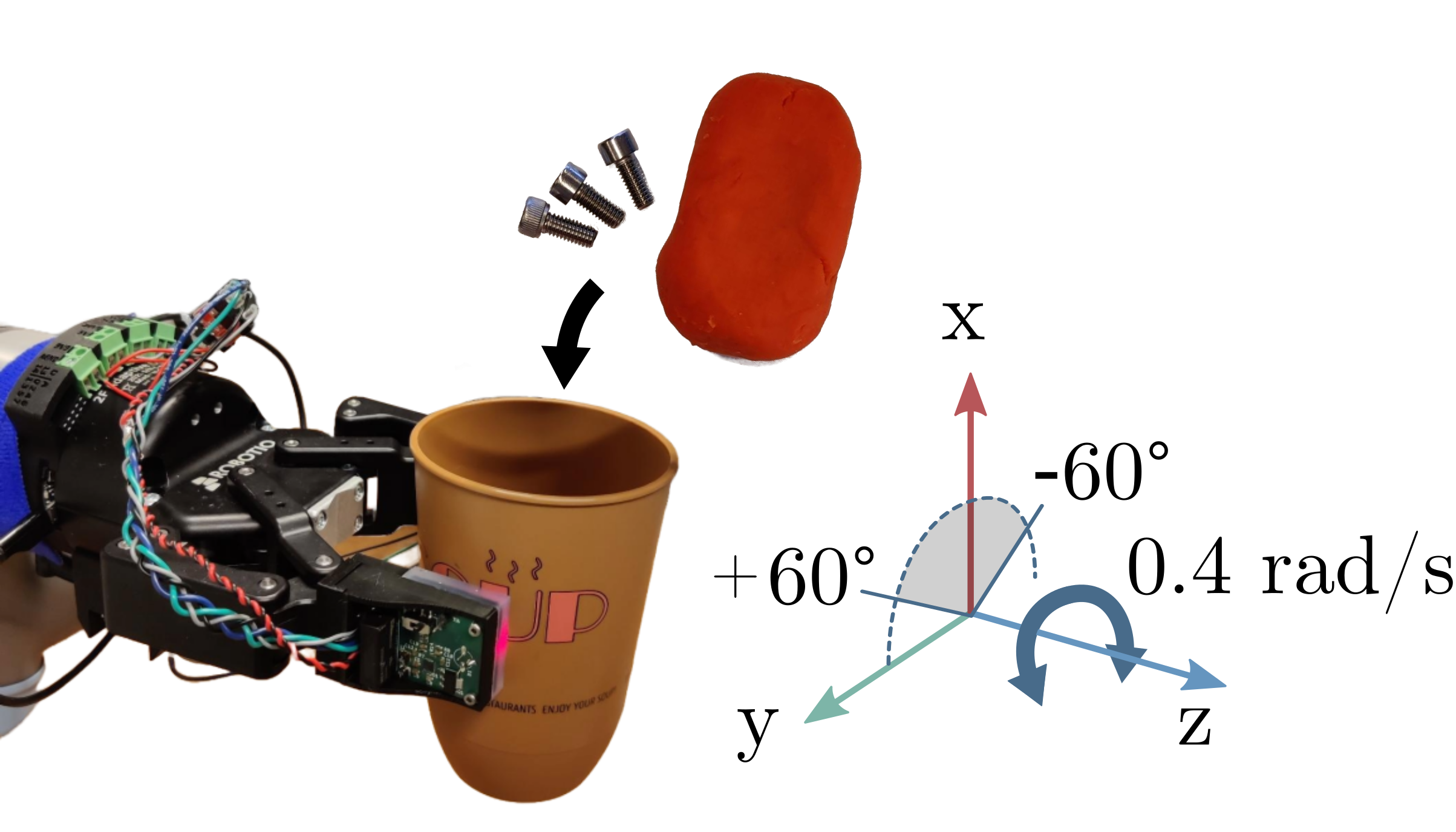}
\caption{Setup for the cup shaking experiment. Either three 14\,mm M6 bolts or a chunk of playdough are placed in the cup while the robot rotates from +60\degree\ to -60\degree.}
\label{fig:cup_shake_setup}
\end{figure}

Fig.~\ref{fig:cup_shake_setup} shows the setup.
The robot is again holding a hard plastic cup.
A single trial consists of the robot turning its wrist from 0\degree\ to an angle of +60\degree, then to -60\degree, and back to 0\degree, all at 0.4\,rad/s.
SMI and acoustic data are recorded simultaneously, at 18\,kHz and 20\,kHz respectively.

We collect 50 trials of the robot holding the empty cup, 50 trials while three 14\,mm M6 bolts are in the cup, and 50 trials with a chunk of playdough in the cup.
These 150 trials comprise the training data for the neural networks. Note that no disturbances are applied to this training dataset.

For the test sets, we first collect four sets of 10 trials each, during which we play sounds next to the robot: white noise, What A Wonderful World by Louis Armstrong, Dark Necessities by Red Hot Chili Peppers, and Bangarang by Skrillex.
In addition, we deliberately try to trick the sensors similarly to Fig.~\ref{fig:bolts_external_cup2}: we manually shake a cup that contains either the bolts or the playdough, while the robot cup is empty. 
This is analogous to multiple robots learning the same task in the same environment.
\enquote{Shaking} entails either mimicking the robot movements, or random, harder shakes. 
For each of these four test scenarios, we collect 10 trials. 
In total, we have 8 test datasets with various disturbances and a single dataset without disturbances that is used for training a classifier to distinguish between the empty cup, a cup containing M6 bolts and a cup containing playdough.

\subsection{Classifier}

For this experiment, we train a separate neural network for each sensor to classify the content of the cup based on the sensor data. As in~\cite{liu2024maniwav}, we use the Audio Spectrogram Transformer (AST)~\cite{gong2021AST}. 
AST is a DeiT transformer~\cite{touvron2021deit} that has been pretrained for audio spectrogram classification, using the Audioset~\cite{audioset} dataset.

Each trial resulted in an 11\,s time series that is resampled to 16\,kHz. 
We use the same pre-processing steps as in AST to convert these into a spectrogram:  We use a log Mel filterbank with 128 bins with a 25\,ms window every 10\,ms. 
This spectrogram is then tokenized using patches that overlap in both the time and frequency dimensions to create the input tokens for the AST transformer.

For each sensor, we train the model using 5-fold cross-validation. 
We train each fold for 10 epochs using a binary cross-entropy loss and select the best checkpoint using the validation score.
To measure performance on the test datasets, we measure the classification accuracy and aggregate the performance on the test dataset over the five models obtained from the different cross-validation splits. This results in an accuracy score and standard deviation for each sensor.

\begin{table}[htb]
\centering
\caption{Classification accuracy for both sensors on the different test datasets.}
\label{tab:classification_accuracy}

\setlength{\tabcolsep}{2pt}
\begin{tabular}{lcc}
\toprule 
\textbf{Experiment}            & \textbf{Microphone  ($\text{acc} \pm \sigma$) } & \textbf{Laser ($\text{acc} \pm \sigma$) } \\ 
\midrule
Validation  (no disturbances) & $1.00 \pm 0.00$ & $ 1.00 \pm 0.00$\\
\midrule 
Ambient sound disturbances & $0.91 \pm 0.10$ & $0.95 \pm 0.03$ \\
\hspace{2pt} Wonderful World - L. A.                & $0.91 \pm 0.03$                     & $0.92 \pm 0.03$                       \\
\hspace{2pt} Dark Necessities - R.\,H.\,C.\,P.    & $0.87 \pm 0.13$                     & $0.95 \pm 0.02$                       \\
\hspace{2pt} Bangarang - Skrillex                 & $0.87 \pm 0.13$                     & $0.96 \pm 0.03$                       \\
\hspace{2pt} White noise              & $1.00 \pm 0.00$                     & $0.97 \pm 0.03$  \\
\midrule 
Targeted sound disturbances & $0.34 \pm 0.42$ & $0.92 \pm 0.15$ \\
\hspace{2pt} Bolts, mimicking robot             & $0.00 \pm 0.00$                     & $0.90 \pm 0.17$                       \\
\hspace{2pt} Bolts, shaking              & $0.00 \pm 0.00$                     & $0.94 \pm 0.13$                       \\
\hspace{2pt} Playdough, mimicking robot        & $1.00 \pm 0.00$                     & $0.86 \pm 0.22$                       \\
\hspace{2pt} Playdough, shaking          & $0.38 \pm 0.08$                     & $1.00 \pm 0.00$                       \\
\bottomrule 
\end{tabular}
\end{table}

\subsection{Results}

\begin{figure}
    \centering
    \begin{subfigure}[t]{0.48\linewidth}
    \vskip 0pt
        \includegraphics[width=0.9\textwidth]{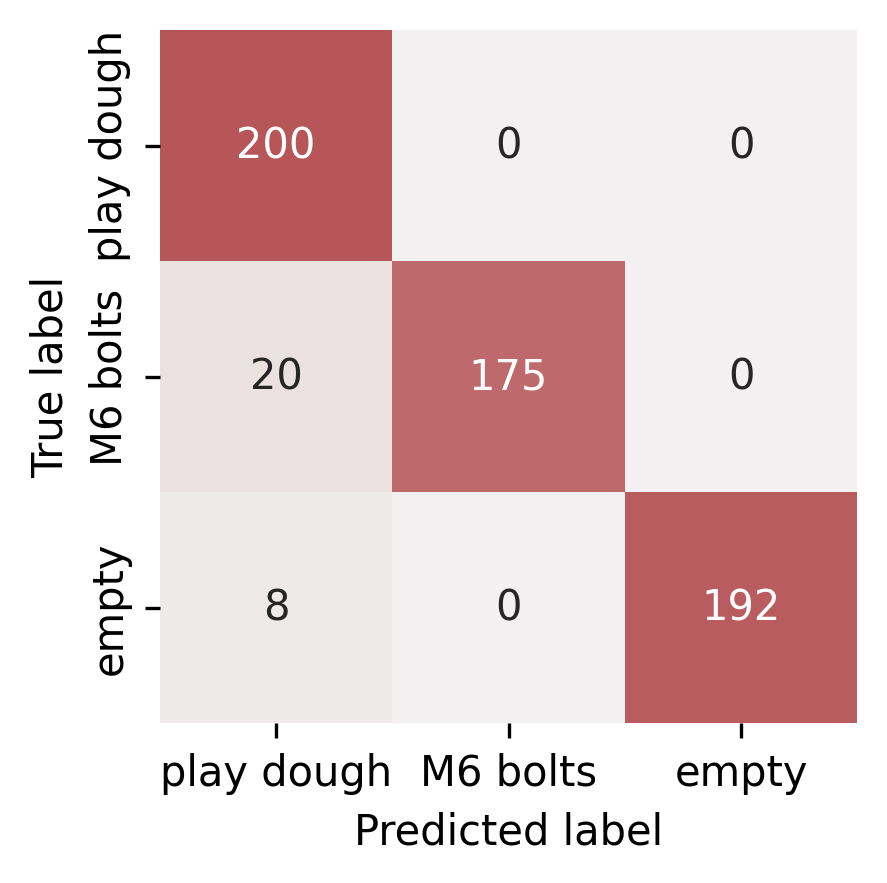}
        \caption{Laser, broadband disturbances.}
        \label{fig:confusion-laser-ambient}
    \end{subfigure}
    \hfill
    \begin{subfigure}[t]{0.48\linewidth}
    \vskip 0pt
        \includegraphics[width=0.9\textwidth]{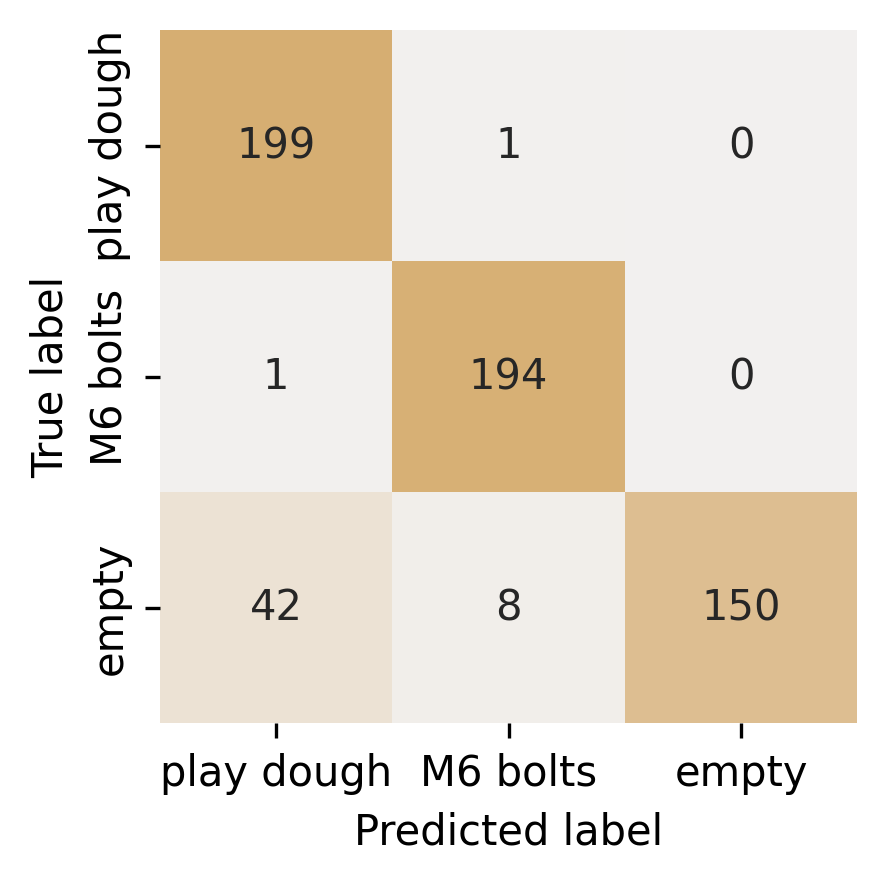}
        \caption{Microphone, broadband disturbances.}
        \label{fig:confusion-mic-ambient}
    \end{subfigure}
    
    \medskip
    
    \begin{subfigure}[t]{0.48\linewidth}
    \vskip 0pt
        \includegraphics[width=0.9\textwidth]{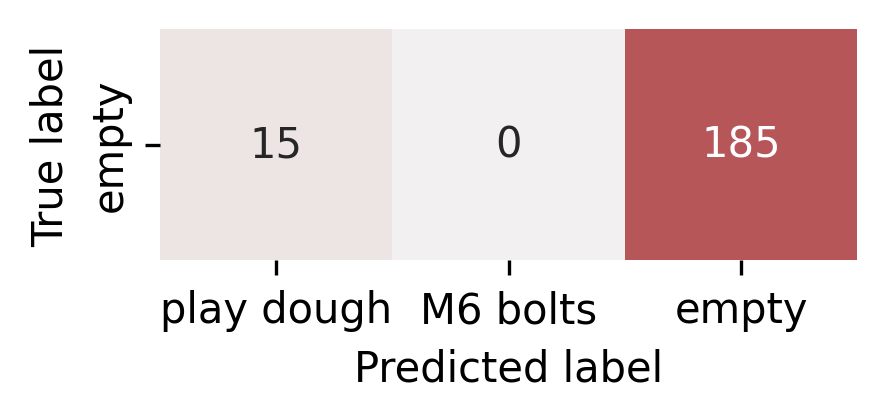}
        \caption{ Laser, targeted disturbances.}
        \label{fig:confusion-laser-targetted}
    \end{subfigure}
    \hfill
    \begin{subfigure}[t]{0.48\linewidth}
    \vskip 0pt
        \includegraphics[width=0.9\textwidth]{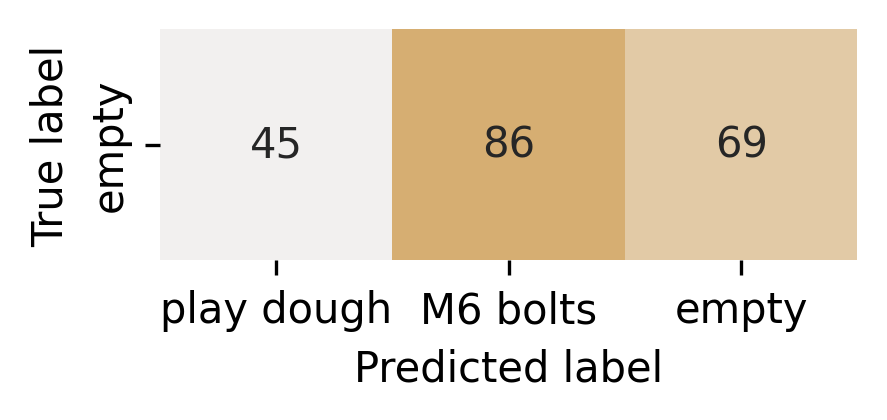}
        \caption{ Microphone, targeted disturbances.}
        \label{fig:confusion-mic-targetted}
    \end{subfigure}
    \caption{Confusion matrices for both sensors for the different sound disturbance experiments.}
    \label{fig:confusion}
\end{figure}

The accuracy scores on the test datasets and the cross-validation score are listed in Table~\ref{tab:classification_accuracy}.
Fig.~\ref{fig:confusion} shows aggregated confusion matrices for ambient and targeted disturbances for each sensor.
Both the laser and the microphone achieve a perfect score on the cross-validation datasets. 
This is consistent with the results from~\cite{liu2024maniwav}, where the authors reported a high success rate (94\,\%) in determining whether the cup was empty.
For the ambient sound disturbances, we find that both sensors experience some deterioration, with the microphone achieving an average accuracy of 91\,\% and the laser achieving 95\,\%. 
For the targeted sound disturbances, the laser has an average accuracy of 92\,\%, clearly outperforming the microphone which only achieves 34\,\%.

\section{Discussion}

The time-domain characterisation indicated increased resilience against ambient noise for SMI, whether it is broadband (Fig.~\ref{fig:silicone_noise_plot}) or mimicking of the desired signal (Fig.~\ref{fig:bolts_external_cup2}).

\begin{figure}[tpb]
\centering
\begin{subfigure}[t]{0.49\textwidth}
    \centering
    \includegraphics[width=\linewidth]{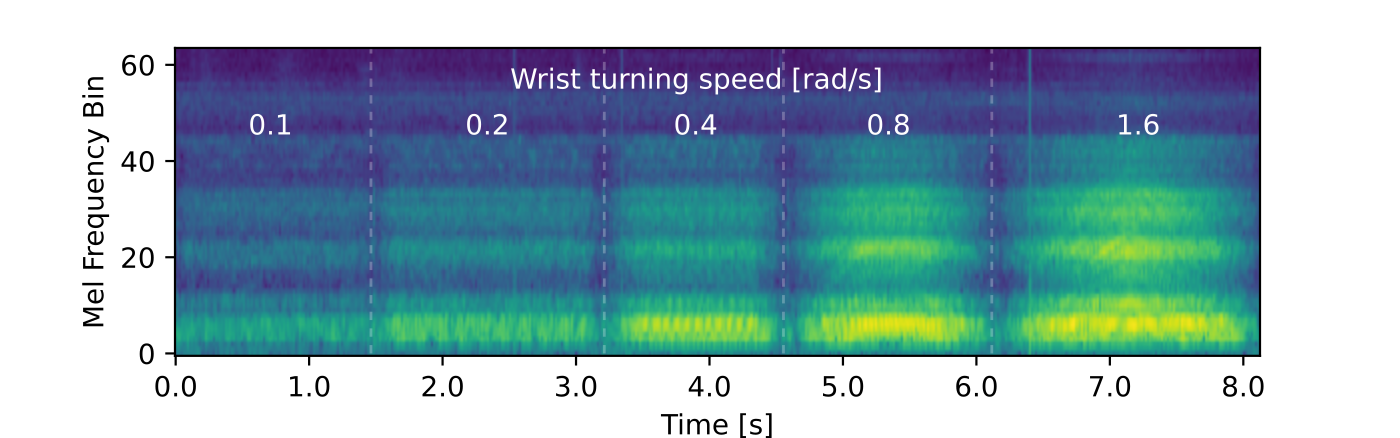}
    \caption{Microphone spectrogram.}
    \label{fig:motor_noise_mic}
\end{subfigure}
\begin{subfigure}[t]{0.49\textwidth}
    \includegraphics[width=\linewidth]{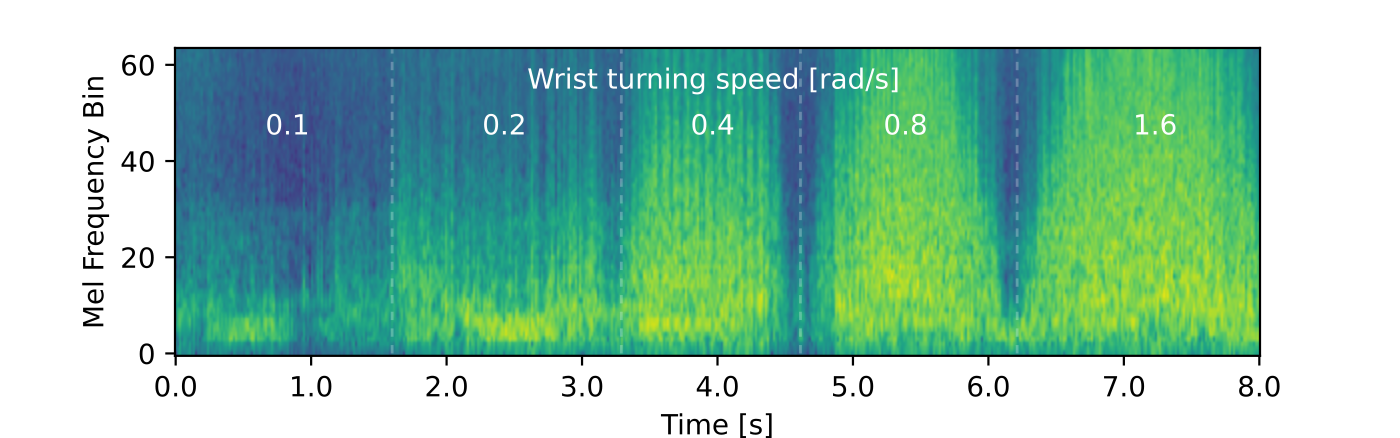} 
    \caption{Laser spectrogram.}
    \label{fig:motor_noise_laser}
\end{subfigure}
\caption{Motor noise at different turning speeds of the robot wrist, plotted with 64 Mel bins (highest frequency: 16\,kHz).}
\label{fig:motor_noise}
\end{figure}

Our frequency domain application showed that the laser did suffer slightly from broadband noise.
The confusion matrix (Fig.~\ref{fig:confusion-laser-ambient}) shows that the dominant error mode for the laser is confusing bolts for playdough.
The microphone performed worse than the laser under broadband noise conditions, but the difference is less than expected given the time-domain analysis (cfr. Fig.~\ref{fig:silicone_noise_plot}).
Even without augmentations in the training data, the AST transformer manages to pick up on the relevant cues in the microphone spectrogram quite well, even when disturbed by broadband noise.
The dominant error mode, see Fig.~\ref{fig:confusion-mic-ambient}, is confusing an empty cup for play dough.
We conclude that the microphone spectrograms are more suited towards distinguishing different events, but that ambient noise can be confused for the more subtle events.
The laser has some difficulty distinguishing between events.
We can understand why this might be the case by plotting the spectrograms when the robot holds an empty cup and shakes at different speeds, see Fig.~\ref{fig:motor_noise}.
Whereas the motor noise presents in clear bands of the microphone spectrogram, it floods the whole spectrum in the laser readout.
Because of this, motor noise can obscure the signal.
Such frequency spreading is fundamental to SMI: the sine wave in Fig.~\ref{fig:smi_principle} is converted to a signal with the same fundamental frequency, but with additional higher frequency components.
In addition, the same frequency spreading effect can occur for the signal itself, meaning that distinguishing features between events could be pushed to frequency bands that are currently not measured.

For the targeted disturbances with bolts, the laser vastly outperforms the microphone, which is consistent with the time-domain experiment of Fig.~\ref{fig:bolts_external_cup2}. 
When mimicking the robot movements with playdough in the cup, the microphone achieves a perfect score because does not pick up soft ambient noises.
When shaking the cup, the sounds are louder, and microphone performance decreases significantly.
Surprisingly, the laser achieves its worst score when mimicking the robot movements with playdough, while still scoring perfectly when shaking with playdough.
We suspect an accidental, uncontrolled disturbance or the non-stationary noise floor described in \cite{proesmans2025selfmixinglaserinterferometryrobotic} to be the cause of this.

\section{Conclusion and Future Work}

In previous work~\cite{proesmans2025selfmixinglaserinterferometryrobotic}, we presented a robotic fingertip making use of SMI for extrinsic
contact sensing as an ambient noise-resilient alternative to
acoustic sensing. 
Here, we extended the time-domain analysis and determined that SMI is not only more robust against broadband noise, but also to ambient sounds mimicking the desired signal.
In the frequency domain, we find that for broadband ambient
noise, SMI still outperforms acoustic sensing, but the difference
is less pronounced than in time-domain analyses. 
For targeted noise disturbances, analogous to multiple robots simultaneously collecting data for the same task, SMI significantly outperforms acoustic sensing.
Lastly, we discussed how motor noise affects SMI sensing more
than acoustic sensing.

In future work, first and foremost the readout frequency of the SMI signal should be increased to confirm if desired signals can be better distinguished from motor noise in higher frequency bands.
This will allow to employ SMI's ambient noise resilience and robust integration capabilities for advanced robotic manipulation.

\balance
\bibliographystyle{IEEEtran}
\bibliography{references.bib}

\end{document}